\documentclass{article} 
\usepackage{iclr2027_conference,times}
\usepackage{graphicx}

\usepackage{amsmath,amsfonts,bm}

\def\eqref#1{equation~\ref{#1}}

\def\1{\bm{1}}

\DeclareMathAlphabet{\mathsfit}{\encodingdefault}{\sfdefault}{m}{sl}
\SetMathAlphabet{\mathsfit}{bold}{\encodingdefault}{\sfdefault}{bx}{n}

\usepackage{float}
\usepackage{hyperref}
\usepackage{url}
\usepackage{enumitem}
\usepackage{xurl}
\usepackage{caption}
\usepackage{wrapfig}
\usepackage{needspace}
\usepackage{makecell}
\usepackage{amsthm}
\newtheorem{proposition}{Proposition}
\usepackage{placeins}
\newcommand{\grouprule}{%
  \noalign{\vskip 2pt\hrule height\arrayrulewidth\vskip 2pt}%
}
\usepackage{xcolor}

\newcommand{\outerrule}{%
  \noalign{\vskip 2pt\hrule height 0.8pt\vskip 2pt}%
}

\usepackage{listings}

\definecolor{promptbackground}{RGB}{248,248,248}
\definecolor{promptborder}{RGB}{205,205,205}
\usepackage{multirow}
\usepackage{amsmath,amssymb,bm}
\usepackage[ruled]{algorithm2e}
\SetAlgoNlRelativeSize{-1}

\SetNlSty{MyNlSty}{}{}

\SetKwFor{ForEach}{\textbf{for}\textnormal{ each}}{\textnormal{do}}{}
\IncMargin{1em}
\DontPrintSemicolon
\SetKwInOut{Input}{Input}
\SetKwInOut{Output}{Output}
\SetKw{Return}{return}

\title{DeKoR: Decomposed Knowledge Routing for Efficient GNN--LLM Fusion}

\author{
Peiyuan Li$^{1,*}$, Yongqi Huang$^{1,*}$, Jitao Zhao$^{1}$, Di Jin$^{1}$ and Dongxiao He$^{1}$ \\
School of Computer Science and Technology \\
Tianjin University, Tianjin, China \\
\texttt{\{lipeiyuan04,yqhuang,zjtao,jindi,hedongxiao\}@tju.edu.cn} \\
$^*$Equal contribution
}

\iclrfinalcopy
\begin{document}

\title{DeKoR: Decomposed Knowledge Routing for Efficient GNN--LLM Fusion}
\maketitle

\fancyhead{} 
\renewcommand{\headrulewidth}{0pt} 

\begin{abstract}
Large Language Models (LLMs) have shown strong potential for text-attributed graph learning by leveraging rich pre-trained knowledge and language understanding capabilities. Combining LLMs with Graph Neural Networks (GNNs) offers a promising paradigm for expanding the modeling capabilities of graph learning. However, LLMs and GNNs follow distinct learning paradigms, leading to a pronounced mismatch between their knowledge spaces. Such discrepancies make effective knowledge fusion nontrivial. Moreover, existing methods often incur substantial additional computation for only limited performance gains. To address these challenges, we propose DeKoR, an efficient framework for GNN and LLM knowledge fusion. To enable efficient knowledge transfer, DeKoR decomposes LLM knowledge into complementary local and global views and uses the LLM as an offline teacher to distill them into lightweight experts. To enable selective knowledge fusion, an adaptive router learns how much each node should rely on each expert. Extensive experiments on multiple benchmarks demonstrate that DeKoR outperforms existing baselines on both node and subgraph classification, achieving superior effectiveness and efficiency.

\end{abstract}

\section{Introduction}
Large Language Models (LLMs) such as DeepSeek-R1~\citep{deepseek-r1} and Qwen3~\citep{qwen3} have demonstrated strong capabilities in understanding and reasoning through advances in pretraining and post-training. These capabilities motivate exploration of how LLM knowledge can support graph learning~\citep{LLM4Graph_survey_1, LLM4Graph_survey_2}. In text-attributed graphs (TAGs), LLMs draw on rich pre-trained knowledge to interpret node texts, while graph models capture dependencies among nodes~\citep{arXiv, TAG_2, TAG_learning, CHoE}. Their integration opens a promising path toward general graph learning, enabling graph models to draw on broader knowledge and support a wider range of application scenarios~\citep{G-retriever, Graph-toolformer}. 


\begin{wrapfigure}[11]{r}{0.47\columnwidth}
    \vspace{-6pt}
    \centering
    \includegraphics[width=\linewidth]{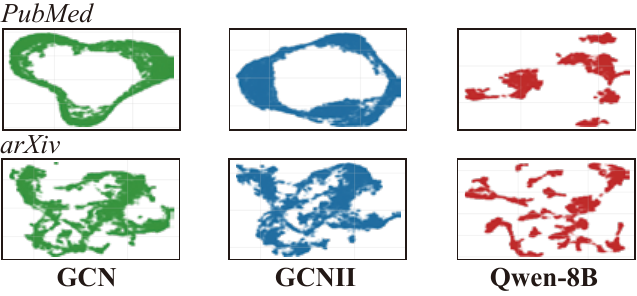}
    \captionsetup{
        justification=raggedright,
        singlelinecheck=false,
        skip=2pt
    }
    \caption{UMAP visualizations of GCN, GCNII, and Qwen-8B on PubMed and arXiv.}
    \label{fig:UMAP}

    \vspace{-5pt}
\end{wrapfigure}

Among methods that use LLMs for graph learning, integrating LLMs with Graph Neural Networks (GNNs) offers a promising approach to TAG learning, termed the GNN--LLM paradigm. By combining the powerful textual understanding of LLMs with the relational modeling of GNNs, these methods can better understand node texts in graph context, capturing richer contextual information. Existing GNN--LLM methods mainly differ in how knowledge is transferred and integrated between the two models. Some approaches integrate semantic knowledge from LLMs into GNNs for subsequent graph learning ~\citep{TAPE, OFA, LinguGKD}; others inject graph structural knowledge into LLMs through graph tokens or parameters~\citep{Graph2Token, GraphLLM, GraphToken, TEA-GLM, GaRA}. Recent studies further explore explicit interactions between structural and semantic knowledge, such as~\citep{ERAlign, GLANCE}. 

Despite these advances, existing GNN--LLM methods still face two challenges in achieving effective and efficient knowledge fusion.
\textbf{Challenge I: Mismatched knowledge space between LLMs and GNNs.} LLMs and GNNs acquire knowledge under fundamentally different learning paradigms, leading to distinct representation spaces. This mismatch is consistently observed across five datasets: as shown in Tab.~\ref{tab:representation_gap}, the representation gaps between GNNs and Qwen-8B~\citep{qwen3}, measured by $1-\mathrm{CKA}$~\citep{CKA}, are consistently larger than those between GCN~\citep{GCN} and GCNII~\citep{GCNII}. The UMAP visualizations~\citep{UMAP} in Fig.~\ref{fig:UMAP} further reveal distinct representation geometries. Such misalignment can cause direct knowledge fusion to introduce redundant or conflicting information, limiting the effective use of the complementary knowledge encoded by the two models.

\begin{wrapfigure}[12]{l}{0.47\columnwidth}
    \vspace{-8pt}
    \centering

    \captionsetup{
        justification=raggedright,
        singlelinecheck=false,
        skip=3pt
    }
    \captionof{table}{Representation gaps among GCN, GCNII, and Qwen-8B across five datasets.}
    \label{tab:representation_gap}

    \setlength{\tabcolsep}{1.5pt}
    \renewcommand{\arraystretch}{0.92}

    \resizebox{\linewidth}{!}{
    \begin{tabular}{lccc}
        \outerrule
        \textbf{Dataset} &
        \makecell{\textbf{GCN}\\[-1pt]\textbf{vs.~GCNII}} &
        \makecell{\textbf{GCN}\\[-1pt]\textbf{vs.~Qwen-8B}} &
        \makecell{\textbf{GCNII}\\[-1pt]\textbf{vs.~Qwen-8B}} \\
        \grouprule
        Cora     & $0.05$ & $0.26$ & $\mathbf{0.29}$ \\
        CiteSeer & $0.05$ & $\mathbf{0.24}$ & $0.21$ \\
        PubMed   & $0.14$ & $\mathbf{0.33}$ & $0.26$ \\
        WikiCS   & $0.19$ & $\mathbf{0.37}$ & $0.30$ \\
        arXiv    & $0.20$ & $\mathbf{0.38}$ & $0.37$ \\
        \grouprule
        Average  & $0.13$ & $\mathbf{0.32}$ & $0.29$ \\
        \outerrule
    \end{tabular}
    }

    \vspace{-4pt}
\end{wrapfigure}

\textbf{Challenge II: Limited gains under high computational cost.}
Many existing GNN--LLM methods require LLM inference for downstream prediction, introducing substantial computational and memory overhead. As shown in Tab.~\ref{tab:efficiency}, several representative methods incur considerably higher inference costs, while their performance gains remain limited. This imbalance restricts the practicality of GNN--LLM integration, particularly in low-resource settings. Effective integration therefore requires preserving useful LLM knowledge without relying on the LLM during inference.

These challenges suggest that effective GNN--LLM integration should avoid directly coupling the original knowledge spaces of the two models. Our key insight is to separate knowledge transfer from knowledge fusion: LLM knowledge is first transformed into lightweight representations, and fusion is then performed after this transfer rather than directly between the two original models. This design avoids directly fusing knowledge from the mismatched GNN and LLM spaces, while allowing the LLM to be removed during inference.

Based on this insight, we propose DeKoR, a knowledge decomposition and routing framework for GNN--LLM fusion. DeKoR decomposes LLM knowledge into two complementary views: local knowledge that preserves fine-grained node information and global knowledge that characterizes each node through its relations to global prototypes. The decomposed knowledge is distilled into lightweight experts, whose predictions are selectively combined by a node-adaptive router. The LLM is used only as an offline teacher, allowing DeKoR to retain useful LLM knowledge without requiring LLM inference during deployment. In summary, our main contributions are as follows:

\begin{itemize}
\item We introduce a new perspective on GNN--LLM knowledge fusion by identifying the mismatch between their knowledge spaces and decoupling knowledge transfer from knowledge fusion. This perspective first transforms LLM knowledge into lightweight forms and then performs fusion on the transferred knowledge.

\item Following this idea, we propose DeKoR. It separates LLM knowledge into complementary local and global views and distills them into lightweight experts. For each node, an adaptive router determines how the two experts contribute to prediction, allowing the model to exploit LLM knowledge without requiring LLM inference.

\item We conduct extensive experiments on multiple benchmarks, demonstrating that DeKoR consistently improves performance across node and subgraph classification tasks while substantially reducing inference time and GPU memory usage compared with most GNN--LLM baselines.
\end{itemize}


\section{Related Work}
\textbf{LLM for Graph Learning.} LLM for Graph Learning aims to leverage the rich pre-trained semantic knowledge of LLMs to assist with graph-related tasks. Methods such as ~\citet{LLaGA, GraphGPT_GET, GraphGPT} investigate how to transform graph data into inputs that can be processed by LLMs. ~\citet{HiGPT, GOFA} focus on employing instruction tuning to enable LLMs to directly handle diverse graph-related tasks. Furthermore, ~\citet{STAGE, OpenGraph} explore how to leverage LLMs to enhance graph data with semantic knowledge. 

\textbf{GNN--LLM Knowledge Fusion.}
 GNN--LLM knowledge fusion methods aim to jointly leverage the structural knowledge captured by GNNs and the semantic knowledge encoded by LLMs for graph learning. One line of work integrates semantic knowledge from LLMs into GNNs. For example, methods such as ~\citet{TAPE, OFA} transform LLM semantic knowledge into representations for subsequent GNN learning, while~\citet{LinguGKD} distills knowledge from an LLM teacher into a GNN through representation alignment. Another line of work injects graph structural knowledge into LLMs. Methods including~\citet{Graph2Token, GraphLLM, GraphToken, TEA-GLM, GaRA} encode structural knowledge into graph tokens or parameters, enabling subsequent LLMs to exploit the information for graph-related tasks. Beyond such directional knowledge transfer, recent studies further explore interactions between structural and semantic knowledge. For instance,~\citet{ERAlign} aligns structural and semantic representations in a shared latent space, whereas~\citet{GLANCE} adaptively determines when LLM knowledge should be incorporated for individual nodes.
 Despite these advances, most existing approaches fuse GNN and LLM knowledge without considering the mismatch between their knowledge spaces. Therefore, their integration may introduce redundant or conflicting information. Moreover, many methods rely on LLM inference, resulting in substantial computational overhead. These limitations motivate a more effective and efficient knowledge integration paradigm.

\section{Preliminary}
\textbf{Text-Attributed Graph.} 
A text-attributed graph (TAG) is denoted as 
$\mathcal{G}=(\mathcal{V},\mathcal{E},\mathbf{X},\mathcal{T},\mathcal{Y})$, where $\mathcal{V} = \{v_1, \ldots, v_{|\mathcal{V}|}\}$ and $\mathcal{E} = \{e_1, \ldots, e_{|\mathcal{E}|}\}$ are the node and edge sets, respectively. Each node $v\in\mathcal{V}$ is associated with a text description $t_v\in\mathcal{T}$, 
which is encoded into a feature vector $x_v\in\mathbf{X}$ by a text encoder. $\mathcal{Y} = \{y_v\}_{v \in \mathcal{V}}$ denotes the set of node labels.

\textbf{Graph Neural Networks.}
GNNs learn node representations by propagating and aggregating information from neighbors. Let
$\mathbf{h}_v^{(0)}=x_v$ denote the initial feature of node $v$, for the
$\ell$-th layer, the representation of node $v$ is updated by
\begin{equation}
\label{eq:1}
\mathbf{h}_v^{(\ell)}
=
\operatorname{UPD}^{(\ell)}
\left(
    \mathbf{h}_v^{(\ell-1)},
    \operatorname{AGG}^{(\ell)}
    \left(
        \left\{
            \mathbf{h}_u^{(\ell-1)} : u \in \mathcal{N}(v)
        \right\}
    \right)
\right),
\end{equation}
where $\mathcal{N}(v)$ denotes the neighbors of node $v$, $\operatorname{AGG}^{(\ell)}(\cdot)$ is an aggregation
function (e.g., mean or sum), and $\operatorname{UPD}^{(\ell)}(\cdot)$ combines the node's previous representation
with the aggregated information.

\textbf{Large Language Models.}
We denote a pre-trained large language model as
$\operatorname{LLM}(\cdot)$. Given a text $t$, the LLM first tokenizes it into a sequence of tokens
$\mathbf{x}=(x_1,x_2,\ldots,x_q)$, which are then mapped to continuous
embeddings and processed through a stack of Transformer layers. Let
$\mathbf{R}^{(L)}=[\mathbf{r}_1,\ldots,\mathbf{r}_q]$ denote the token-level
representations produced by the last Transformer layer, where
$\mathbf{r}_i\in\mathbb{R}^{d_l}$ corresponds to token $x_i$.
We extract node-level semantic representations from $\mathbf{R}^{(L)}$ using a representation extraction operation
$\operatorname{Rep}(\cdot)$:
\begin{equation}
\label{eq:2}
\mathbf{z}
=
\operatorname{LLM}(t)
=
\operatorname{Rep}\!\left(\mathbf{R}^{(L)}\right)
\in\mathbb{R}^{d_z},
\end{equation}
where $\operatorname{Rep}(\cdot)$ can be implemented using mean pooling or the last valid token, and $d_z$ denotes the dimension of the resulting semantic representation.
In this work,  we construct textual prompts for each node $v$ with text $t_v$, and use the pre-trained LLM to extract its semantic representation.

\section{Method}

\begin{figure}[t]
    \centering
    \includegraphics[
        width=\columnwidth,
        keepaspectratio
    ]{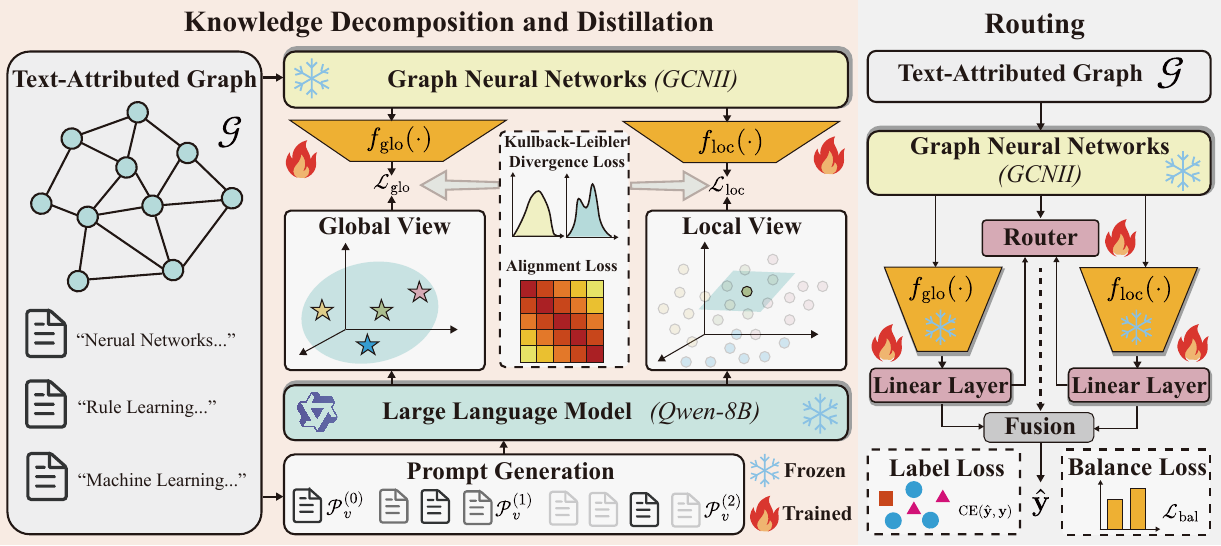}
    \caption{Overall framework of DeKoR. LLM semantic knowledge is decomposed into complementary global and local views and distilled into lightweight experts. During routing, a node-adaptive router dynamically combines the two experts for final prediction.}
    \label{fig:model}
\end{figure}

We propose DeKoR, an efficient framework for GNN--LLM knowledge fusion, as illustrated in Fig.~\ref{fig:model}. DeKoR first constructs structural and semantic representations. It then decomposes the LLM knowledge into local and global semantic views, which capture node-level semantics and prototype-based global relations, respectively. The two forms of knowledge are distilled into lightweight semantic experts, without requiring LLM inference during deployment. Finally, a node-adaptive router assigns node-specific weights to the two experts and combines their predictions.

\subsection{Representation Construction}
Given a text-attributed graph
$\mathcal{G}=(\mathcal{V},\mathcal{E},\mathbf{X},\mathcal{T},\mathcal{Y})$,
we pre-train a GNN encoder $f_{\mathrm{GNN}}(\cdot)$ and fine-tune an $\operatorname{LLM}(\cdot)$ for each
dataset. Both models are then frozen and used
to construct structural and semantic representations for each
node. Detailed training procedures and settings are
provided in Appendix~\ref{app:gnn_pretraining} and Appendix~\ref{app:llm_finetuning}.

We adopt the frozen $f_{\mathrm{GNN}}(\cdot)$ to generate structural representations. For each node
$v\in\mathcal{V}$, its structural representation is given by $\mathbf{h}_v = f_{\mathrm{GNN}}(\mathbf{X}, \mathcal{E})_v$. To obtain semantic representations, we construct three textual
prompts for each node at different context scales: (i) the center
node text only, (ii) the center node text with sampled 1-hop
neighbors, and (iii) the center node text with sampled 2-hop
neighbors. We denote these prompts as
$\mathcal{P}_v^{(0)},\ \mathcal{P}_v^{(1)},\ \mathcal{P}_v^{(2)}$,
respectively. The detailed prompt templates are provided in
Appendix~\ref{app:llm_prompting}. We then use the frozen
$\operatorname{LLM}(\cdot)$ to extract semantic embeddings from
these prompts:
\begin{equation}
\label{eq:3}
\mathbf{e}_v^{(m)} = \operatorname{LLM}(\mathcal{P}_v^{(m)}) \in\mathbb{R}^{d_z},
\quad m\in\{0,1,2\}.
\end{equation}
Finally, we concatenate the three embeddings to obtain the semantic
representation:
\begin{equation}
\label{eq:4}
\mathbf{z}_v =
[{\mathbf{e}}_v^{(0)};{\mathbf{e}}_v^{(1)};
{\mathbf{e}}_v^{(2)}]
\in \mathbb{R}^{3d_z}.
\end{equation}

\subsection{LLM Semantic Knowledge Decomposition}

\begin{proposition}[Multi-Granularity Semantic Representations]
\label{pro:multi_granularity_semantics}
Let $\mathbf{z}\in\mathbb{R}^{d}$ denote a node semantic
representation, and let $\mathbf{C}\in\mathbb{R}^{K\times d}$ denote a set of $K$ semantic
prototypes. We define the prototype-induced representation as
\begin{equation}
\label{eq:5}
\mathbf{q}(\mathbf{z})
=
\operatorname{softmax}
\left(
\frac{\mathbf{C}\mathbf{z}}
{\tau}
\right),
\end{equation}
where $\tau>0$ is a temperature parameter.
The Jacobian~\citep{Jacobian} of this mapping satisfies
\begin{equation}
\label{eq:6}
\operatorname{rank}
\left(
\mathbf{J}_{\mathbf{q}}(\mathbf{z})
\right)
\leq
\min
\left\{
K-1,\operatorname{rank}(\mathbf{C})
\right\}.
\end{equation}
Moreover, for any perturbation
$\boldsymbol{\delta}\in\mathbb{R}^{d}$ satisfying
$\mathbf{C}\boldsymbol{\delta}=a\mathbf{1}$
for some scalar $a$, the prototype-induced representation remains
invariant: $\mathbf{q}(\mathbf{z}+\boldsymbol{\delta})=\mathbf{q}(\mathbf{z}).$

Therefore, $\mathbf{z}$ and $\mathbf{q}(\mathbf{z})$ represent semantics at different granularities: $\mathbf{z}$ preserves original semantics, while $\mathbf{q}(\mathbf{z})$ captures the node's relationships with semantic prototypes. The proof is provided in Appendix~\ref{app:proof_prop1}.
\end{proposition}

\paragraph{Connection to Our Method.}
Proposition~\ref{pro:multi_granularity_semantics} characterizes how to decompose semantic knowledge at different granularities. In DeKoR, we preserve the original LLM representations $\mathbf{Z}$ as fine-grained local semantic knowledge, while constructing $\mathbf{Q}$ as prototype-induced global semantic knowledge. 

Specifically, we apply $K$-means~\citep{K-means} to
$\{\mathbf{z}_v\}_{v\in\mathcal{V}}$ and obtain $K$ global semantic
prototypes
$\mathbf{C}=[\mathbf{c}_1,\ldots,\mathbf{c}_K]^\top$.
For each node $v$, we compute its similarity with these prototypes as:
\begin{equation}
\label{eq:7}
\mathbf{q}_v
=
\operatorname{softmax}
\left(
\frac{{\mathbf{z}}_v\mathbf{C}^{\top}}
{\tau_{\mathrm{teacher}}}
\right),
\end{equation}
where $\tau_{\mathrm{teacher}}$ controls the sharpness of the distribution. We collect the resulting
representations as:
\begin{equation}
\label{eq:8}
\mathbf{Q}
=
[\mathbf{q}_1,\mathbf{q}_2,\ldots,\mathbf{q}_n]^\top
\in\mathbb{R}^{n\times K}.
\end{equation}
Accordingly, we decompose LLM semantic knowledge into complementary local and global views, capturing semantic information at different granularities.

\subsection{Local-Global Semantic Knowledge Distillation}

After decomposing the LLM knowledge into local and global semantic views, we transfer the knowledge into lightweight modules through knowledge distillation. We provide a theoretical characterization of the knowledge preserved by the distillation objectives.

\begin{proposition}[Information Preservation under Knowledge Distillation]
\label{pro: Information Preservation in Knowledge Distillation}
Let $\mathbf{Z}$ and $\mathbf{H}^{\mathrm{loc}}$ denote the teacher
representation and its distilled representation from the local module, respectively. Under the standard InfoNCE assumption~\citep{InfoNCE}, we have
\begin{equation}
\label{eq:9}
I\!\left(
\mathbf{H}^{\mathrm{loc}};\mathbf{Z}
\right)
\geq
\log |\mathcal{B}|
-
\mathcal{L}_{\mathrm{nce}},
\end{equation}
where $\mathcal{B}$ denotes the training batch. Thus, minimizing
$\mathcal{L}_{\mathrm{nce}}$ increases the lower bound on the mutual
information retained by the lightweight module. For the global semantic
view, let $\mathbf{q}_v$ and $\mathbf{p}_v^{\mathrm{glo}}$ denote
the teacher and distilled prototype distributions from the global module, respectively. If
$\mathcal{L}_{\mathrm{glo},v}
=
\mathrm{KL}(
\mathbf{q}_v\|
\mathbf{p}_v^{\mathrm{glo}}
)$, Pinsker's inequality~\citep{Pinsker} gives
\begin{equation}
\label{eq:10}
\left\|
\mathbf{q}_v-\mathbf{p}_v^{\mathrm{glo}}
\right\|_1
\leq
\sqrt{
2\mathcal{L}_{\mathrm{glo},v}
}.
\end{equation}
Therefore, the two objectives support the preservation of local and global semantic knowledge, respectively.
The proof is provided in Appendix~\ref{app:proof_prop2}.
\end{proposition}

\paragraph{Connection to Our Method.}
Proposition~\ref{pro: Information Preservation in Knowledge Distillation}
supports distilling the decomposed LLM knowledge into lightweight
modules. The InfoNCE objective preserves local semantic knowledge, while the KL objective preserves the global prototype distribution.

We define lightweight local and global semantic experts as $f_{\mathrm{loc}}(\cdot)$ and $f_{\mathrm{glo}}(\cdot)$. Given the structural representations $\mathbf{H}$ generated by the GNN, we use $f_{\mathrm{loc}}(\cdot)$ and $f_{\mathrm{glo}}(\cdot)$ to encode $\mathbf{H}$ into local and global semantic representations:
\begin{equation}
\label{eq:11}
{\mathbf{H}}^{\mathrm{loc}} = f_{\mathrm{loc}}(\mathbf{H}),
\qquad
{\mathbf{H}}^{\mathrm{glo}} = f_{\mathrm{glo}}(\mathbf{H}).
\end{equation}
For the local semantic expert, we encourage $\mathbf{h}_v^{\mathrm{loc}} \in \mathbf{H}^{\mathrm{loc}}$ to align with $\mathbf{z}_v \in \mathbf{Z}$ through cosine similarity. For a batch $\mathcal{B}$, the cosine similarity loss $\mathcal{L}_{\mathrm{cos}}$ is defined as:
\begin{equation}
\label{eq:12}
\mathcal{L}_{\mathrm{cos}}
=
1-\frac{1}{|\mathcal{B}|}
\sum_{v\in\mathcal{B}}
\cos(\mathbf{h}_v^{\mathrm{loc}},\mathbf{z}_v).
\end{equation}
Consistent with Proposition~\ref{pro: Information Preservation in Knowledge Distillation}, we introduce a symmetric InfoNCE loss. The directional loss $\mathcal{L}_{\mathrm{h}\rightarrow\mathrm{z}}$ is defined from $\mathbf{h}_v^{\mathrm{loc}}$ to $\mathbf{z}_v$, while
$\mathcal{L}_{\mathrm{z}\rightarrow\mathrm{h}}$ is obtained by reversing the two representations:
\begin{equation}
\label{eq:13}
\begin{aligned}
\mathcal{L}_{\mathrm{h}\rightarrow\mathrm{z}}
&=
-\frac{1}{|\mathcal{B}|}
\sum_{v\in\mathcal{B}}
\log
\frac{
\exp\left(
\cos(\mathbf{h}_v^{\mathrm{loc}},\mathbf{z}_v)/\tau_{\mathrm{loc}}
\right)
}{
\sum_{u\in\mathcal{B}}
\exp\left(
\cos(\mathbf{h}_v^{\mathrm{loc}},\mathbf{z}_u)/\tau_{\mathrm{loc}}
\right)
},
\qquad
\mathcal{L}_{\mathrm{nce}}
=
\frac{1}{2}
\left(
\mathcal{L}_{\mathrm{h}\rightarrow\mathrm{z}}
+
\mathcal{L}_{\mathrm{z}\rightarrow\mathrm{h}}
\right),
\end{aligned}
\end{equation}
where $\tau_{\mathrm{loc}}$ is the temperature parameter. The local expert is optimized with:
\begin{equation}
\label{eq:14}
\mathcal{L}_{\mathrm{loc}}=\mathcal{L}_{\mathrm{cos}}+\lambda_{\mathrm{nce}}\mathcal{L}_{\mathrm{nce}}.
\end{equation}
For the global semantic expert, we align $\mathbf{h}_v^{\mathrm{glo}} \in \mathbf{H}^{\mathrm{glo}}$ with $\mathbf{q}_v \in \mathbf{Q}$. We first convert $\mathbf{h}_v^{\mathrm{glo}}$ into a probability distribution using softmax. Motivated by Proposition~\ref{pro: Information Preservation in Knowledge Distillation}, we then measure its discrepancy from $\mathbf{q}_v$ using the KL divergence~\citep{KL}:
\begin{equation}
\label{eq:15}
\mathcal{L}_{\mathrm{glo}}
=
\mathrm{KL}\!\left(
\mathbf{q}_v
\;\middle\|\;
\operatorname{softmax}\!\left(
\frac{\mathbf{h}_v^{\mathrm{glo}}}{\tau_{\mathrm{glo}}}
\right)
\right),
\end{equation}
where $\tau_{\mathrm{glo}}$ is the temperature parameter. Finally, the parameters of $f_{\mathrm{loc}}(\cdot)$ and $f_{\mathrm{glo}}(\cdot)$ are updated by minimizing
$\mathcal{L}_{\mathrm{distill}}
=
\lambda_{\mathrm{loc}}\mathcal{L}_{\mathrm{loc}}
+
\lambda_{\mathrm{glo}}\mathcal{L}_{\mathrm{glo}}$.

\subsection{Node-Adaptive Semantic Knowledge Routing}
After distillation, we freeze $f_{\mathrm{loc}}(\cdot)$ and $f_{\mathrm{glo}}(\cdot)$. Given the $\mathbf{h}_v^{\mathrm{loc}}$ and $\mathbf{h}_v^{\mathrm{glo}}$ for each node $v\in\mathcal{V}$, we use prediction heads $g^{\mathrm{loc}}(\cdot)$ and $g^{\mathrm{glo}}(\cdot)$ to obtain their prediction logits:
\begin{equation}
\label{eq:16}
\mathbf{y}_v^{\mathrm{loc}} = g^{\mathrm{loc}}(\mathbf{h}_v^{\mathrm{loc}}),
\qquad
\mathbf{y}_v^{\mathrm{glo}} = g^{\mathrm{glo}}\!\left(
\mathbf{h}_v^{\mathrm{glo}}
\right).
\end{equation}

To achieve fine-grained knowledge fusion, we train a router $\phi(\cdot)$ to adaptively assign fusion weights to the two experts for each node:
\begin{equation}
\label{eq:17}
\boldsymbol{\alpha}_v
=
\operatorname{softmax}\!\left(
\phi\!\left(
\left[
\mathbf{h}_v;
\mathbf{y}_v^{\mathrm{loc}};
\mathbf{y}_v^{\mathrm{glo}}
\right]
\right)
\right)
=
\left[
\alpha_v^{\mathrm{loc}},
\alpha_v^{\mathrm{glo}}
\right]^\top,
\end{equation}
where $\boldsymbol{\alpha}_v$ denotes the fusion weights for node $v$. Then we can obtain the final prediction $\hat{\mathbf{y}}_v$:
\begin{equation}
\label{eq:18}
\hat{\mathbf{y}}_v
=
\alpha_v^{\mathrm{loc}}\mathbf{y}_v^{\mathrm{loc}}
+
\alpha_v^{\mathrm{glo}}\mathbf{y}_v^{\mathrm{glo}}.
\end{equation}
To encourage balanced utilization of the experts during routing, we define the average routing weight of expert $m$ in a batch as $\bar{\alpha}^{m}=\frac{1}{|\mathcal{B}|}\sum_{v\in\mathcal{B}}\alpha_v^{m}$, and the balance loss $\mathcal{L}_{\mathrm{bal}}$ is defined as:
\begin{equation}
\label{eq:19}
\mathcal{L}_{\mathrm{bal}}
=
\sum_{m\in\{\mathrm{loc},\mathrm{glo}\}}
\left(
\bar{\alpha}^{m}-\frac{1}{2}
\right)^2.
\end{equation}
At this stage, we optimize the parameters of the router $\phi(\cdot)$, the
prediction $g^{\mathrm{loc}}(\cdot)$ and $g^{\mathrm{glo}}(\cdot)$ by minimizing the routing objective $\mathcal{L}_{\mathrm{route}}=\mathrm{CE}(\hat{\mathbf{y}}_v,y_v)+\lambda_{\mathrm{bal}}\mathcal{L}_{\mathrm{bal}}.$
\section{Experiments}

In this section, we conduct extensive experiments to evaluate the effectiveness of \textbf{DeKoR},
aiming to answer the following research questions:
\textbf{RQ1:} How effective is DeKoR compared with existing baselines on node and subgraph classification tasks? 
\textbf{RQ2:} Does DeKoR remain effective across different GNN and LLM backbones? 
\textbf{RQ3:} How do the proposed components contribute to the performance of DeKoR? 
\textbf{RQ4:} How does the adaptive routing mechanism behave across different nodes?
\textbf{RQ5:} How sensitive is DeKoR to key hyper-parameters? 
\textbf{RQ6:} How efficient is DeKoR compared with existing baselines?

\subsection{Experimental setup}

\textbf{Datasets.} In node and subgraph classification experiments, five widely
used text-attributed graph datasets are employed, including Cora~\citep{CoraCiteseer}, CiteSeer~\citep{CoraCiteseer}, PubMed~\citep{Pubmed}, WikiCS~\citep{Wikics}, and arXiv~\citep{arXiv}. For details on the utilized datasets, refer to Appendix~\ref{app:node_dataset} and \ref{app:subgraph_dataset}.  

\textbf{Baselines.} The baseline models include four typical GNNs: GCN~\citep{GCN}, GAT~\citep{GAT}, GraphSAGE~\citep{GraphSAGE}, and GIN~\citep{GIN}; four graph Transformer models (GTs): DIFFFormer~\citep{DIFFormer}, NodeFormer~\citep{NodeFormer}, DUALFormer~\citep{DUALFormer}, and CoBFormer~\citep{CoBFormer}; and five GNN--LLM methods: UniGraph~\citep{UniGraph}, OFA~\citep{OFA}, GraphGPT~\citep{GraphGPT}, LinguGKD~\citep{LinguGKD}, and GLANCE~\citep{GLANCE}. Appendix~\ref{app:baselines} provides detailed descriptions of the compared baselines.

\subsection{Experimental results}

\begin{table*}[h]
\caption{Accuracy(\%) of the node classification task across five datasets. Best and runner-up methods are \textbf{bolded} and \underline{underlined}, respectively.}
\label{tab:node_classification}
\vspace{\baselineskip}
\centering
\setlength{\tabcolsep}{5pt}
\renewcommand{\arraystretch}{1.05}
\resizebox{\textwidth}{!}{
\begin{tabular}{clccccc}
\outerrule
\textbf{Category} & \textbf{Method} & \textbf{Cora} & \textbf{CiteSeer} & \textbf{PubMed} & \textbf{WikiCS} & \textbf{arXiv} \\
\grouprule
\multirow{4}{*}{GNNs}
& \texttt{GCN}\hspace{0.35em}{\tiny\textcolor{gray}{\textit{ICLR'17}}} & $87.40 \pm 0.96$ & $74.95 \pm 1.38$ & $88.70 \pm 0.35$ & $82.25 \pm 0.36$ & $66.57 \pm 0.45$ \\
& \texttt{GAT}\hspace{0.35em}{\tiny\textcolor{gray}{\textit{ICLR'18}}} & $87.99 \pm 0.47$ & $75.83 \pm 0.96$ & $87.51 \pm 0.77$ & $82.71 \pm 0.67$ & $65.64 \pm 0.99$ \\
& \texttt{GraphSAGE}\hspace{0.35em}{\tiny\textcolor{gray}{\textit{NeurIPS'17}}} & $88.53 \pm 1.26$ & $76.87 \pm 1.34$ & $90.60 \pm 0.43$ & $84.01 \pm 0.45$ & $68.40 \pm 0.63$ \\
& \texttt{GIN}\hspace{0.35em}{\tiny\textcolor{gray}{\textit{ICLR'19}}} & $86.46 \pm 0.87$ & $73.65 \pm 1.32$ & $84.85 \pm 1.06$ & $61.08 \pm 1.51$ & $55.05 \pm 2.70$ \\
\grouprule
\multirow{4}{*}{GTs}
& \texttt{DIFFormer}\hspace{0.35em}{\tiny\textcolor{gray}{\textit{ICLR'23}}} & $87.15 \pm 1.41$ & $77.12 \pm 0.69$ & $91.37 \pm 0.27$ & $83.29 \pm 0.72$ & $70.43 \pm 0.21$ \\
& \texttt{NodeFormer}\hspace{0.35em}{\tiny\textcolor{gray}{\textit{NeurIPS'22}}} & $86.71 \pm 1.82$ & $76.91 \pm 1.03$ & $91.12 \pm 0.24$ & $80.44 \pm 0.29$ & $48.25 \pm 0.85$ \\
& \texttt{DUALFormer}\hspace{0.35em}{\tiny\textcolor{gray}{\textit{ICLR'25}}} & $88.43 \pm 0.60$ & $74.95 \pm 1.01$ & $87.17 \pm 0.88$ & $63.24 \pm 1.08$ & $54.57 \pm 0.56$ \\
& \texttt{CoBFormer}\hspace{0.35em}{\tiny\textcolor{gray}{\textit{ICML'24}}} & $84.59 \pm 1.91$ & $75.07 \pm 0.69$ & $89.98 \pm 1.12$ & $79.03 \pm 0.75$ & $68.69 \pm 0.38$ \\
\grouprule
\multirow{5}{*}{\shortstack{LLM for\\Graphs}}
& \texttt{UniGraph}\hspace{0.35em}{\tiny\textcolor{gray}{\textit{KDD'25}}} & $82.32 \pm 2.01$ & $73.48 \pm 2.01$ & $76.59 \pm 1.19$ & $80.60 \pm 0.89$ & $51.19 \pm 2.10$ \\
& \texttt{OFA}\hspace{0.35em}{\tiny\textcolor{gray}{\textit{ICLR'24}}} & $87.49 \pm 1.09$ & $74.49 \pm 0.31$ & $89.66 \pm 0.39$ & $80.52 \pm 0.74$ & $67.97 \pm 0.65$ \\
& \texttt{GraphGPT}\hspace{0.35em}{\tiny\textcolor{gray}{\textit{SIGIR'24}}} & $81.60 \pm 0.76$ & $65.29 \pm 2.43$ & $65.94 \pm 0.90$ & $73.67 \pm 1.26$ & $54.49 \pm 0.58$ \\
& \texttt{LinguGKD}\hspace{0.35em}{\tiny\textcolor{gray}{\textit{AAAI'25}}} & $87.91 \pm 1.23$ & $75.70 \pm 0.36$ & $85.05 \pm 0.13$ & $82.26 \pm 0.22$ & $62.70 \pm 2.65$ \\
& \texttt{GLANCE}\hspace{0.35em}{\tiny\textcolor{gray}{\textit{ICLR'26}}} & $\underline{89.50 \pm 0.40}$ & $\underline{80.72 \pm 2.39}$ & $\underline{92.60 \pm 0.10}$ & $\mathbf{88.30 \pm 2.51}$ & $\underline{71.79 \pm 1.08}$ \\
\grouprule
\multirow{1}{*}{Ours}
& \texttt{DeKoR} & $\mathbf{90.79 \pm 0.34}$ & $\mathbf{81.47 \pm 1.83}$ & $\mathbf{93.13 \pm 0.73}$ & $\underline{88.22 \pm 1.85}$ & $\mathbf{75.30 \pm 1.15}$ \\
\outerrule
\end{tabular}%
}
\end{table*}

\textbf{Node Classification (RQ1).} 
Tab.~\ref{tab:node_classification} reports the node classification results. DeKoR achieves the highest accuracy on four of the five datasets and achieves competitive performance with the best-performing baseline on WikiCS, demonstrating the effectiveness of our method. DeKoR achieves an average accuracy of 85.78\% across the five datasets, exceeding the strongest baseline, GLANCE (84.58\%), by 1.20\%. The largest improvement occurs on arXiv, where DeKoR achieves 75.30\% accuracy, 3.51\% above GLANCE. These results show that DeKoR performs consistently across the evaluated datasets, including the large-scale datasets.
\begin{table*}[t]
\caption{Accuracy(\%) of the subgraph classification task across five datasets. Best and runner-up methods are \textbf{bolded} and \underline{underlined}, respectively.}
\label{tab:graph_classification}
\vspace{\baselineskip}
\centering
\setlength{\tabcolsep}{5pt}
\renewcommand{\arraystretch}{1.05}
\resizebox{\textwidth}{!}{
\begin{tabular}{clccccc}
\outerrule
\textbf{Category} & \textbf{Method} & \textbf{Cora} & \textbf{CiteSeer} & \textbf{PubMed} & \textbf{WikiCS} & \textbf{arXiv} \\
\grouprule
\multirow{4}{*}{GNNs}
& \texttt{GCN}\hspace{0.35em}{\tiny\textcolor{gray}{\textit{ICLR'17}}} & $87.30 \pm 0.59$ & $74.74 \pm 0.84$ & $84.14 \pm 0.75$ & $79.37 \pm 0.93$ & $67.67 \pm 0.49$ \\
& \texttt{GAT}\hspace{0.35em}{\tiny\textcolor{gray}{\textit{ICLR'18}}} & $86.95 \pm 0.45$ & $74.24 \pm 1.47$ & $84.08 \pm 0.92$ & $79.06 \pm 0.85$ & $67.75 \pm 0.56$ \\
& \texttt{GraphSAGE}\hspace{0.35em}{\tiny\textcolor{gray}{\textit{NeurIPS'17}}} & $87.25 \pm 0.60$ & $74.99 \pm 1.17$ & $83.77 \pm 0.68$ & $79.49 \pm 0.55$ & $68.20 \pm 0.42$ \\
& \texttt{GIN}\hspace{0.35em}{\tiny\textcolor{gray}{\textit{ICLR'19}}} & $86.46 \pm 1.07$ & $74.49 \pm 0.69$ & $84.17 \pm 0.38$ & $78.23 \pm 0.98$ & $66.20 \pm 0.31$ \\
\grouprule
\multirow{4}{*}{GTs}
& \texttt{DIFFormer}\hspace{0.35em}{\tiny\textcolor{gray}{\textit{ICLR'23}}} & $87.30 \pm 0.77$ & $75.74 \pm 1.70$ & $84.10 \pm 0.49$ & $79.36 \pm 0.72$ & $67.82 \pm 0.61$ \\
& \texttt{NodeFormer}\hspace{0.35em}{\tiny\textcolor{gray}{\textit{NeurIPS'22}}} & $87.54 \pm 0.39$& $73.99 \pm 0.92$ & $83.76 \pm 1.06$ & $75.47 \pm 1.93$ & $66.33 \pm 0.20$ \\
& \texttt{DUALFormer}\hspace{0.35em}{\tiny\textcolor{gray}{\textit{ICLR'25}}} & $87.44 \pm 0.53$ & $75.83 \pm 0.97$ & $84.35 \pm 1.01$ & $79.54 \pm 0.82$ & $68.81 \pm 0.44$ \\
& \texttt{CoBFormer}\hspace{0.35em}{\tiny\textcolor{gray}{\textit{ICML'24}}} & $86.46 \pm 0.84$ & $75.45 \pm 1.14$ & $83.96 \pm 1.24$ & $78.53 \pm 1.17$ & $66.23 \pm 0.84$ \\
\grouprule
\multirow{5}{*}{\shortstack{LLM for\\Graphs}}
& \texttt{UniGraph}\hspace{0.35em}{\tiny\textcolor{gray}{\textit{KDD'25}}} & $80.40 \pm 0.87$ & $71.94 \pm 1.27$ & $75.02 \pm 0.81$ & $72.56 \pm 1.15$ & $46.17 \pm 3.17$ \\
& \texttt{OFA}\hspace{0.35em}{\tiny\textcolor{gray}{\textit{ICLR'24}}} & $86.76 \pm 1.26$ & $75.41 \pm 0.98$ & $\mathbf{89.61 \pm 0.36}$ & $\underline{79.70 \pm 0.93}$ & $67.95 \pm 0.73$ \\
& \texttt{GraphGPT}\hspace{0.35em}{\tiny\textcolor{gray}{\textit{SIGIR'24}}} & $52.98 \pm 1.96$ & $42.49 \pm 1.09$ & $75.97 \pm 0.69$ & $44.77 \pm 1.27$ & $52.86 \pm 2.49$ \\
& \texttt{LinguGKD}\hspace{0.35em}{\tiny\textcolor{gray}{\textit{AAAI'25}}} & $\underline{87.64 \pm 0.37}$ & $77.08 \pm 1.45$ & $83.52 \pm 0.41$ & $78.94 \pm 1.50$ & $63.92 \pm 0.52$ \\
& \texttt{GLANCE}\hspace{0.35em}{\tiny\textcolor{gray}{\textit{ICLR'26}}} & $87.10 \pm 1.03$ & $\underline{77.24 \pm 1.76}$ & $84.69 \pm 1.16$ & $79.30 \pm 1.06$ & $\underline{69.42 \pm 0.46}$ \\
\grouprule
Ours
& \texttt{DeKoR} & $\mathbf{88.23 \pm 0.51}$ & $\mathbf{78.00 \pm 1.64}$ & $\underline{85.11 \pm 0.77}$ & $\mathbf{80.23 \pm 0.88}$ & $\mathbf{70.10 \pm 0.10}$ \\
\outerrule
\end{tabular}%
}
\end{table*}

\textbf{Subgraph Classification (RQ1).} To further evaluate the effectiveness of our proposed method, we conduct subgraph classification experiments. The subgraph construction is described in Appendix~\ref{app:subgraph_dataset}. As shown in Tab.~\ref{tab:graph_classification}, DeKoR outperforms all baseline models on four of the five datasets, achieving the best accuracy on Cora, CiteSeer, WikiCS, and arXiv. In particular, it achieves 78.00\% on CiteSeer, outperforming the runner-up by 0.76\% points. On PubMed, OFA achieves the highest accuracy of 89.61\%, while DeKoR reaches 85.11\%. This may be attributed to OFA's unified task formulation, which explicitly incorporates graph-level tasks into its model design. Overall, these results demonstrate the effectiveness of DeKoR on the subgraph classification task.

\begin{table*}[h]
\caption{Accuracy(\%) of DeKoR with different GNN and LLM backbones on node classification.}
\label{tab:backbone}
\scriptsize
\setlength{\tabcolsep}{3.5pt}
\renewcommand{\arraystretch}{1.05}

\noindent
\begin{minipage}[t]{0.34\textwidth}
    \centering
    \textbf{(a) GNN Backbones}\par

    \begin{tabular*}{0.90\linewidth}
        {@{\hspace{5pt}}l@{\extracolsep{\fill}}c@{\hspace{5pt}}}
    \outerrule
    \textbf{Backbone} & \textbf{Avg. Accuracy} \\
    \grouprule
    \texttt{GCN} & $82.09 \pm 1.25$ \\
    \texttt{GraphSAGE} & $78.02 \pm 4.53$ \\
    \texttt{GCNII} & $82.86 \pm 1.22$ \\
    \grouprule
    \texttt{GCN (Ours)} & $83.99 \pm 1.46$ \\
    \texttt{GraphSAGE (Ours)} & $85.97 \pm 2.05$ \\
    \texttt{GCNII (Ours)} & $85.78 \pm 1.18$ \\
    \outerrule
    \end{tabular*}
\end{minipage}%
\hspace{0.04\textwidth}%
\begin{minipage}[t]{0.62\textwidth}
    \centering
    \textbf{(b) LLM Backbones}\par
    \begin{tabular*}{0.92\linewidth}
        {@{\hspace{5pt}}l@{\extracolsep{\fill}}ccc@{\hspace{5pt}}}
    \outerrule
    \textbf{Backbone} & \textbf{Cora} & \textbf{CiteSeer} & \textbf{WikiCS} \\
    \grouprule
    \texttt{Llama-8B} & $90.00 \pm 0.98$ & $78.77 \pm 4.53$ & $85.95 \pm 1.93$ \\
    \texttt{Vicuna-7B} & $88.58 \pm 2.36$ & $79.69 \pm 6.03$ & $84.69 \pm 2.06$ \\
    \texttt{Qwen-8B} & $86.56 \pm 4.23$ & $79.17 \pm 3.54$ & $82.21 \pm 1.73$ \\
    \grouprule
    \texttt{Llama-8B (Ours)} & $90.59\pm0.68$ & $80.97\pm2.94$ & $87.99\pm2.06$ \\
    \texttt{Vicuna-7B (Ours)} & $89.91\pm0.74$ & $80.89\pm2.82$ & $87.92\pm1.91$ \\
    \texttt{Qwen-8B (Ours)} & $90.79 \pm 0.34$ & $81.47 \pm 1.83$ & $88.22 \pm 1.85$ \\
    \outerrule
    \end{tabular*}
\end{minipage}
\end{table*}
\textbf{Backbone Analysis (RQ2).}
As shown in Tab.~\ref{tab:backbone}, we evaluate DeKoR across different GNN and LLM backbones. In Tab.~\ref{tab:backbone}(a), we replace the GNN backbone and report the average accuracy across five datasets. Compared with directly using the GNN backbone for classification, DeKoR increases the average accuracy of GCN, GraphSAGE, and GCNII by 1.90\%, 7.95\%, and 2.92\%, respectively. Moreover, the performance of DeKoR remains stable across different GNN backbones, ranging from 83.99\% to 85.97\%. In Tab.~\ref{tab:backbone}(b), DeKoR improves accuracy over each corresponding LLM-only baseline on all three datasets. Take Qwen-8B as an example, accuracy rises from 86.56\% to 90.79\% on Cora, from 79.17\% to 81.47\% on CiteSeer, and from 82.21\% to 88.22\% on WikiCS. These results suggest that DeKoR effectively coordinates the complementary knowledge from GNNs and LLMs while maintaining robustness to changes in backbones.

\begin{figure}[t]
    \centering
    \includegraphics[
        width=\textwidth
    ]{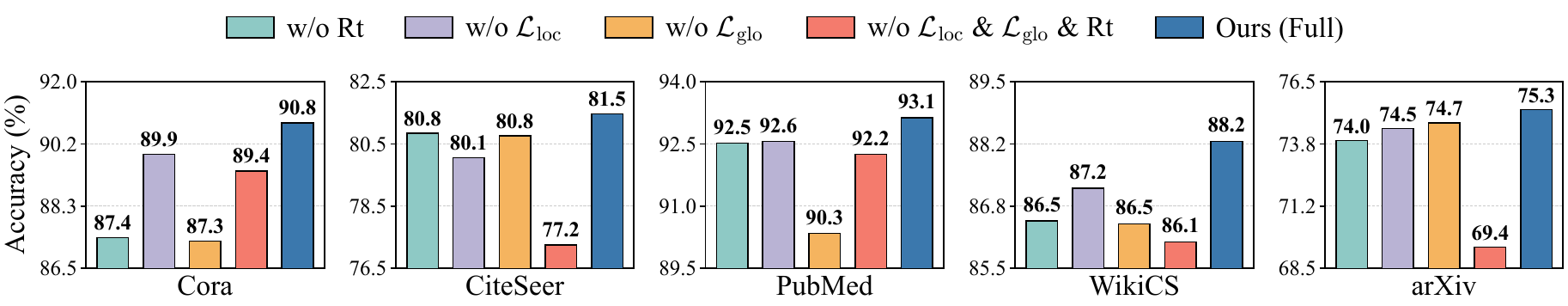}
    \caption{Ablation studies of key components in DeKoR.}
    \label{fig:ablation}
\end{figure}

\textbf{Ablation Study (RQ3).} To evaluate the contribution of key components of DeKoR, we conduct an ablation study. We design four variants: (1) w/o Rt, which removes the routing mechanism; (2) w/o $\mathcal{L}_{\mathrm{loc}}$, which removes the local distillation loss; (3) w/o $\mathcal{L}_{\mathrm{glo}}$, which removes the global distillation loss; and (4) w/o $\mathcal{L}_{\mathrm{loc}}$ \& $\mathcal{L}_{\mathrm{glo}}$ \& Rt, which removes both distillation losses and the routing mechanism. As shown in Fig.~\ref{fig:ablation}, removing any component leads to a performance drop across all datasets, demonstrating their contributions to DeKoR. Specifically, the average accuracy is reduced by 1.54\%, 0.92\% and 1.86\% by removing the router, the local distillation loss or the global distillation loss. Removing these three components will produce the largest average decline of 2.92\%.

\begin{wrapfigure}[16]{l}{0.45\columnwidth}
    \vspace{-13pt}
    \centering
    \includegraphics[
        width=\linewidth
    ]{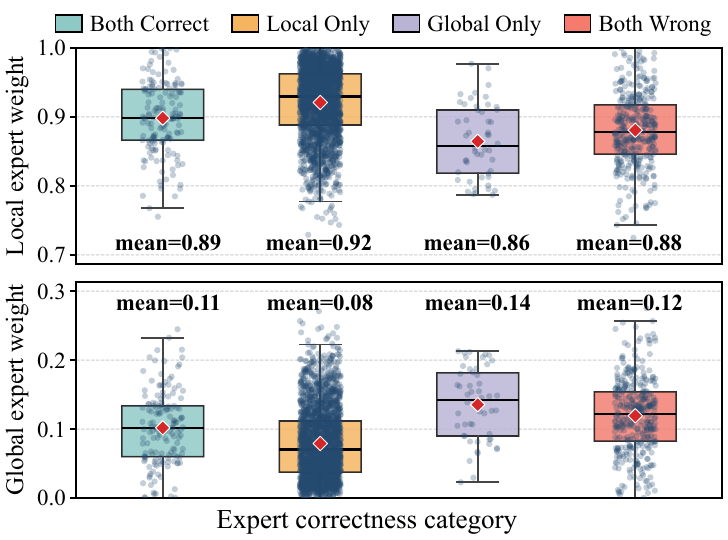}
    \caption{Routing weights of local and global semantic experts grouped by expert classification correctness on WikiCS.}
    \label{fig:router-correctness-wikics}
    \vspace{0pt}
\end{wrapfigure}

\textbf{Routing Analysis (RQ4).} We further analyze the routing behavior of the proposed method. We partition the nodes into four groups based on the classification correctness. Fig.~\ref{fig:router-correctness-wikics} shows the distributions of the routing weights for these groups on WikiCS, where individual points represent nodes, boxes indicate the interquartile ranges, and red diamonds denote the mean weights.
The local semantic expert receives a higher routing weight than the global semantic expert across all groups, suggesting that local semantic knowledge plays a more prominent role on WikiCS. Moreover, the router favors the local semantic expert in the
\emph{Local Only} group but gives more weight to the global semantic
expert in the \emph{Global Only} group. This suggests that the router assigns higher weights to experts with correct predictions.

\begin{table*}[t]
\caption{Inference time (ms) and GPU memory usage (MiB) across five datasets.}
\label{tab:efficiency}
\vspace{\baselineskip}
\centering
\setlength{\tabcolsep}{6pt}
\renewcommand{\arraystretch}{1.05}

\resizebox{\textwidth}{!}{
\begin{tabular}{lc|cccccc}
\outerrule
\textbf{Method} & \textbf{Time/Memory}
& \textbf{Cora} & \textbf{CiteSeer} & \textbf{PubMed}
& \textbf{WikiCS} & \textbf{arXiv} & \textbf{Avg. Accuracy(\%)} \\
\grouprule

\multirow{2}{*}{\texttt{UniGraph}}
& Time
& $1363594.3$ & $680603.5$ & $775438.3$ & $1172503.5$ & $2426704.3$ &  \multirow{2}{*}{$72.84$} \\
& Memory
& $14363$ & $14368$ & $14338$ & $14367$ & $14596$ & \\

\grouprule

\multirow{2}{*}{\texttt{OFA}}
& Time
& $55285.8$ & $54932.2$ & $69428.5$ & $70333.5$ & $65201.4$ &  \multirow{2}{*}{$80.03$} \\
& Memory
& $13027$ & $13027$ & $13028$ & $13026$ & $13027$ & \\

\grouprule

\multirow{2}{*}{\texttt{GraphGPT}}
& Time
& $768910.8$ & $495775.1$ & $471579.6$ & $1548980.4$ & $5966955.7$ &  \multirow{2}{*}{$68.20$} \\
& Memory
& $15351$ & $15359$ & $16213$ & $16253$ & $15821$ & \\

\grouprule

\multirow{2}{*}{\texttt{LinguGKD}}
& Time
& $3.5$ & $2.4$ & $4.4$ & $2.3$ & $1.7$ &  \multirow{2}{*}{$78.72$} \\
& Memory
& $163$ & $226$ & $123$ & $145$ & $34$ & \\

\grouprule

\multirow{2}{*}{\texttt{GLANCE}}
& Time
& $21433.0$ & $20641.1$ & $21709.5$ & $271763.1$ & $27176.3$ &  \multirow{2}{*}{$\underline{84.58}$} \\
& Memory
& $22911$ & $22911$ & $22911$ & $22911$ & $22912$ & \\

\grouprule

\multirow{2}{*}{\texttt{DeKoR}}
& Time
& $1041.4$ & $854.1$ & $801.0$ & $965.8$ & $810.0$ &  \multirow{2}{*}{$\mathbf{85.78}$} \\
& Memory
& $60$ & $60$ & $60$ & $60$ & $62$ & \\

\outerrule
\end{tabular}%
}
\end{table*}

\begin{figure}[h]
    \centering
    \includegraphics[
        width=\columnwidth,
        keepaspectratio
    ]{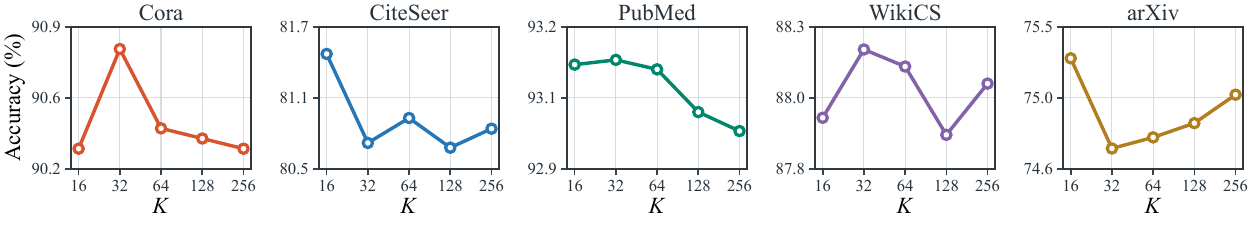}
    \caption{Sensitivity analysis of the number of prototypes $K$.}
    \label{fig:hyper-parameters}
\end{figure}

\textbf{Hyper-parameter Sensitivity Analysis (RQ5).} As shown in Fig.~\ref{fig:hyper-parameters}, we investigate the sensitivity of DeKoR to the number of global semantic prototypes $K$. We can observe that our method maintains stable performance across different values of $K \in \{16, 32, 64, 128, 256\}$, with performance variations within 0.7\% across all five datasets. This indicates low sensitivity to the number of prototypes, with $K=32$ or $K=16$ already providing competitive performance.

\textbf{Efficiency Analysis (RQ6).}
To evaluate the time and memory efficiency of DeKoR, we compare its inference time and GPU memory usage with GNN--LLM baselines across five datasets. We randomly select 1,000 nodes from each dataset for evaluation. As shown in Tab.~\ref{tab:efficiency}, DeKoR requires substantially less GPU memory than most GNN--LLM baselines. Moreover, its inference time is also considerably lower than that of GNN--LLM methods, except LinguGKD. Although LinguGKD achieves lower inference time, it requires more GPU memory than DeKoR on four of the five datasets and achieves 7.06\% lower average accuracy. These results demonstrate that DeKoR substantially reduces the inference-time and memory overhead of GNN--LLM methods while maintaining strong predictive performance.


\section{Conclusion}
In this work, we study how to integrate GNN and LLM knowledge without directly coupling their mismatched knowledge spaces or relying on costly LLM inference. We propose DeKoR, which separates knowledge transfer from knowledge fusion by first reorganizing LLM knowledge before integrating it with graph models. Specifically, DeKoR decomposes LLM knowledge into complementary local and global views, distills them into lightweight experts, and uses a node-adaptive router to determine their contributions to individual predictions. This design allows useful LLM knowledge to be retained while removing the LLM from downstream inference. Extensive experiments on node and subgraph classification benchmarks show consistent improvements in both predictive performance and computational efficiency. These results demonstrate that transferring and reorganizing LLM knowledge before fusion provides a practical way to combine the complementary capabilities of GNNs and LLMs.

\subsection*{AI use statement}

In this work, we used generative AI tools to improve the clarity, organization, and readability of the manuscript and to support literature retrieval and discovery. All AI-assisted suggestions were
manually reviewed and verified by three authors against the original experiments and cited literature. Generative AI was not used to generate experimental results or make scientific conclusions. We take full responsibility for the final content of this work.

\bibliography{iclr2027_conference}
\bibliographystyle{iclr2027_conference}

\clearpage
\appendix

\section{Proofs of Theoretical Results}
\label{app:proofs}

\raggedbottom

\subsection{Proof of Proposition 1}
\label{app:proof_prop1}

\paragraph{Proof.}
Let
$\mathbf{z}\in\mathbb{R}^{d}$
denote a node's semantic representation and let
$\mathbf{C}\in\mathbb{R}^{K\times d}$
contain the $K$ semantic prototypes as its rows. For notational
convenience, we treat $\mathbf{z}$ as a column vector and define
\begin{equation}
\mathbf{r}
=
\frac{\mathbf{C}\mathbf{z}}{\tau},
\qquad
\mathbf{q}
=
\operatorname{softmax}(\mathbf{r}),
\end{equation}
where
$\mathbf{q}\in\mathbb{R}^{K}$ satisfies
$q_k\geq 0$ and
$\sum_{k=1}^{K}q_k=1$.

To examine the local behavior of the prototype-based mapping, we consider the Jacobian of the softmax function with respect to $\mathbf{r}$
\begin{equation}
\frac{\partial \mathbf{q}}
{\partial \mathbf{r}}
=
\operatorname{Diag}(\mathbf{q})
-
\mathbf{q}\mathbf{q}^{\top}.
\label{eq:softmax_jacobian}
\end{equation}
Since
$\mathbf{r}=\mathbf{C}\mathbf{z}/\tau$,
the chain rule gives
\begin{equation}
\mathbf{J}_{\mathbf{q}}(\mathbf{z})
=
\frac{1}{\tau}
\left(
\operatorname{Diag}(\mathbf{q})
-
\mathbf{q}\mathbf{q}^{\top}
\right)
\mathbf{C}.
\label{eq:prototype_jacobian}
\end{equation}

Because
$\mathbf{q}^{\top}\mathbf{1}=1$, we have
\begin{equation}
\left(
\operatorname{Diag}(\mathbf{q})
-
\mathbf{q}\mathbf{q}^{\top}
\right)
\mathbf{1}
=
\mathbf{q}
-
\mathbf{q}
\left(
\mathbf{q}^{\top}\mathbf{1}
\right)
=
\mathbf{0}.
\end{equation}
Therefore,
$\operatorname{Diag}(\mathbf{q})
-\mathbf{q}\mathbf{q}^{\top}$
has rank at most $K-1$. Using
$\operatorname{rank}(\mathbf{A}\mathbf{B})
\leq
\min\{\operatorname{rank}(\mathbf{A}),
\operatorname{rank}(\mathbf{B})\}$,
Eq.~\eqref{eq:prototype_jacobian} yields
\begin{equation}
\operatorname{rank}
\left(
\mathbf{J}_{\mathbf{q}}(\mathbf{z})
\right)
\leq
\min
\left\{
K-1,
\operatorname{rank}(\mathbf{C})
\right\}.
\end{equation}
Hence, the prototype-based representation is sensitive to a
restricted set of directions defined by the prototype geometry,
rather than preserving every variation in the original representation $\mathbf{z}$.

We next establish the invariance property. Consider an arbitrary perturbation
$\boldsymbol{\delta}\in\mathbb{R}^{d}$ satisfying $\mathbf{C}\boldsymbol{\delta}=a\mathbf{1}$
for some scalar $a$. The prototype logits of the perturbed representation become
\begin{equation}
\frac{
\mathbf{C}(\mathbf{z}+\boldsymbol{\delta})
}{\tau}
=
\frac{\mathbf{C}\mathbf{z}}{\tau}
+
\frac{a}{\tau}\mathbf{1}.
\end{equation}
The softmax function is invariant to adding the same scalar to all
logits, i.e.,
\begin{equation}
\operatorname{softmax}
\left(
\mathbf{x}+b\mathbf{1}
\right)
=
\operatorname{softmax}(\mathbf{x})
\end{equation}
for any vector $\mathbf{x}$ and scalar $b$. Therefore,
\begin{equation}
\begin{aligned}
\mathbf{q}(\mathbf{z}+\boldsymbol{\delta})
&=
\operatorname{softmax}
\left(
\frac{
\mathbf{C}(\mathbf{z}+\boldsymbol{\delta})
}{\tau}
\right)\\
&=
\operatorname{softmax}
\left(
\frac{\mathbf{C}\mathbf{z}}{\tau}
+
\frac{a}{\tau}\mathbf{1}
\right)\\
&=
\operatorname{softmax}
\left(
\frac{\mathbf{C}\mathbf{z}}{\tau}
\right)
=
\mathbf{q}(\mathbf{z}).
\end{aligned}
\end{equation}

This invariance shows that the prototype-induced representation
discards variations that affect all prototype similarities equally,
while retaining variations that modify their relative relationships.
Consequently, the original representation and the prototype-based
representation characterize semantic knowledge at different
granularities: the former preserves fine-grained node-specific
semantics, whereas the latter emphasizes prototype-relative global
semantic structure. This completes the proof.
\hfill$\square$\par

\subsection{Proof of Proposition 2}
\label{app:proof_prop2}

\paragraph{Proof.}
We prove the information-preservation properties of the local and
global knowledge separately.

\textbf{Local semantic view.}
Let the random variables
$\mathbf{H}^{\mathrm{loc}}$ and $\mathbf{Z}$ denote the local semantic
representation produced by the lightweight modules and the corresponding
LLM teacher representation, respectively. Consider a batch
$\mathcal{B}$ of $|\mathcal{B}|$ positive pairs
$(\mathbf{h}_v^{\mathrm{loc}},\mathbf{z}_v)$ sampled from their joint
distribution
$p(\mathbf{H}^{\mathrm{loc}},\mathbf{Z})$.
For each anchor $\mathbf{h}_v^{\mathrm{loc}}$, the representations
$\mathbf{z}_u$, $u\neq v$, are treated as negative samples drawn from
the marginal distribution of $\mathbf{Z}$.

Using the similarity critic
\begin{equation}
s(
\mathbf{h}^{\mathrm{loc}},
\mathbf{z}
)
=
\frac{
\cos(
\mathbf{h}^{\mathrm{loc}},
\mathbf{z}
)
}{
\tau_{\mathrm{loc}}
},
\end{equation}
the directional InfoNCE objective is
\begin{equation}
\mathcal{L}_{\mathrm{h}\rightarrow\mathrm{z}}
=
-\mathbb{E}
\left[
\log
\frac{
\exp
s(
\mathbf{h}_v^{\mathrm{loc}},
\mathbf{z}_v
)
}{
\sum_{u\in\mathcal{B}}
\exp
s(
\mathbf{h}_v^{\mathrm{loc}},
\mathbf{z}_u
)
}
\right].
\end{equation}
Under the standard InfoNCE sampling construction, the InfoNCE
lower bound on mutual information gives
\begin{equation}
I\!\left(
\mathbf{H}^{\mathrm{loc}};
\mathbf{Z}
\right)
\geq
\log |\mathcal{B}|
-
\mathcal{L}_{\mathrm{h}\rightarrow\mathrm{z}}.
\label{eq:infonce_hz_bound}
\end{equation}

By exchanging the roles of
$\mathbf{H}^{\mathrm{loc}}$ and $\mathbf{Z}$, the reverse-direction
objective similarly satisfies
\begin{equation}
I\!\left(
\mathbf{Z};
\mathbf{H}^{\mathrm{loc}}
\right)
\geq
\log |\mathcal{B}|
-
\mathcal{L}_{\mathrm{z}\rightarrow\mathrm{h}}.
\label{eq:infonce_zh_bound}
\end{equation}
Since mutual information is symmetric,
\begin{equation}
I\!\left(
\mathbf{H}^{\mathrm{loc}};
\mathbf{Z}
\right)
=
I\!\left(
\mathbf{Z};
\mathbf{H}^{\mathrm{loc}}
\right).
\end{equation}
Therefore, Eqs.~\eqref{eq:infonce_hz_bound} and
\eqref{eq:infonce_zh_bound} imply
\begin{equation}
\begin{aligned}
I\!\left(
\mathbf{H}^{\mathrm{loc}};
\mathbf{Z}
\right)
&\geq
\max
\left\{
\log |\mathcal{B}|
-
\mathcal{L}_{\mathrm{h}\rightarrow\mathrm{z}},
\right.\\[-2pt]
&\hspace{28mm}\left.
\log |\mathcal{B}|
-
\mathcal{L}_{\mathrm{z}\rightarrow\mathrm{h}}
\right\}.
\end{aligned}
\end{equation}
Since the maximum of two values is no smaller than their average,
we further obtain
\begin{equation}
\begin{aligned}
I\!\left(
\mathbf{H}^{\mathrm{loc}};
\mathbf{Z}
\right)
&\geq
\log |\mathcal{B}|
-
\frac{1}{2}
\left(
\mathcal{L}_{\mathrm{h}\rightarrow\mathrm{z}}
+
\mathcal{L}_{\mathrm{z}\rightarrow\mathrm{h}}
\right)\\
&=
\log |\mathcal{B}|
-
\mathcal{L}_{\mathrm{nce}}.
\end{aligned}
\label{eq:symmetric_infonce_bound}
\end{equation}
Hence, minimizing the symmetric InfoNCE objective increases its
corresponding lower bound on the mutual information shared between the
local semantic expert and the LLM representation.

\textbf{Global semantic view.}
For node $v$, let
$\mathbf{q}_v$ denote the prototype-association distribution
from the LLM teacher representation, and define the distribution
predicted by the global semantic module as
\begin{equation}
\mathbf{p}_v^{\mathrm{glo}}
=
\operatorname{softmax}\!\left(
\frac{
\mathbf{h}_v^{\mathrm{glo}}
}{
\tau_{\mathrm{glo}}
}
\right).
\end{equation}
Both $\mathbf{q}_v$ and $\mathbf{p}_v^{\mathrm{glo}}$ are probability
distributions over the $K$ semantic prototypes. The node-wise global
distillation loss is
\begin{equation}
\mathcal{L}_{\mathrm{glo},v}
=
\mathrm{KL}\!\left(
\mathbf{q}_v
\;\middle\|\;
\mathbf{p}_v^{\mathrm{glo}}
\right).
\end{equation}

By Pinsker's inequality, the total variation distance between two
probability distributions satisfies
\begin{equation}
D_{\mathrm{TV}}
\left(
\mathbf{q}_v,
\mathbf{p}_v^{\mathrm{glo}}
\right)
\leq
\sqrt{
\frac{1}{2}
\mathrm{KL}\!\left(
\mathbf{q}_v
\;\middle\|\;
\mathbf{p}_v^{\mathrm{glo}}
\right)
}.
\end{equation}
Using
\begin{equation}
D_{\mathrm{TV}}
\left(
\mathbf{q}_v,
\mathbf{p}_v^{\mathrm{glo}}
\right)
=
\frac{1}{2}
\left\|
\mathbf{q}_v
-
\mathbf{p}_v^{\mathrm{glo}}
\right\|_1,
\end{equation}
we obtain
\begin{equation}
\left\|
\mathbf{q}_v
-
\mathbf{p}_v^{\mathrm{glo}}
\right\|_1
\leq
\sqrt{
2
\mathrm{KL}\!\left(
\mathbf{q}_v
\;\middle\|\;
\mathbf{p}_v^{\mathrm{glo}}
\right)
}
=
\sqrt{
2\mathcal{L}_{\mathrm{glo},v}
}.
\label{eq:global_kl_bound}
\end{equation}
Therefore, decreasing the KL distillation loss directly tightens an
upper bound on the discrepancy between the prototype-association
distribution from global semantic knowledge and the
distribution reconstructed by the lightweight global expert. In
particular, as
$\mathcal{L}_{\mathrm{glo},v}\rightarrow 0$, we have
\begin{equation}
\left\|
\mathbf{q}_v
-
\mathbf{p}_v^{\mathrm{glo}}
\right\|_1
\rightarrow 0.
\end{equation}

Together, Eq.~\eqref{eq:symmetric_infonce_bound} and
Eq.~\eqref{eq:global_kl_bound} show that the two distillation objectives
provide complementary preservation guarantees for the decomposed
knowledge: the local objective preserves shared semantic
information with the original LLM representation, whereas the global
objective preserves the prototype-based semantic distribution. This
completes the proof.
\hfill$\square$\par

\section{Algorithm Description}
The overall procedure of DeKoR is summarized in Algorithm~\ref{alg:ours}.

\begin{algorithm}[h]
\small
\SetAlgoSkip{smallskip}
\caption{DeKoR}
\label{alg:ours}

\Input{
Text-attributed graph $\mathcal{G}$;
pre-trained $f_{\mathrm{GNN}}$ and $\operatorname{LLM}$;
number of prototypes $K$
}

\Output{Predictions $\hat{\mathbf{y}}$}

Obtain structural representations
$\mathbf{H}$ using $f_{\mathrm{GNN}}$\;

Construct prompts and obtain LLM representations
$\mathbf{Z}$ as in~\eqref{eq:3}--\eqref{eq:4}\;

Cluster $\mathbf{Z}$ into $K$ semantic prototypes
and construct global knowledge $\mathbf{Q}$
as in~\eqref{eq:7}--\eqref{eq:8}\;

Initialize semantic experts
$f_{\mathrm{loc}}$ and $f_{\mathrm{glo}}$\;

\While{distillation not converged}{
    Compute
    $\mathbf{H}^{\mathrm{loc}}$ and
    $\mathbf{H}^{\mathrm{glo}}$
    as in~\eqref{eq:11}\;

    Compute local and global distillation losses
    as in~\eqref{eq:12}--\eqref{eq:15}\;

    Update $f_{\mathrm{loc}}$ and $f_{\mathrm{glo}}$
    with $\mathcal{L}_{\mathrm{distill}}$\;
}

Freeze $f_{\mathrm{loc}}$ and $f_{\mathrm{glo}}$\;

\While{routing not converged}{
    Obtain expert predictions as in~\eqref{eq:16}\;

    Compute node-adaptive routing weights
    and fused predictions as in~\eqref{eq:17}--\eqref{eq:18}\;

    Compute $\mathcal{L}_{\mathrm{route}}$
    with the balance loss in ~\eqref{eq:19}\;

    Update $\phi$, $g^{\mathrm{loc}}$, and $g^{\mathrm{glo}}$\;
}

\Return{$\hat{\mathbf{y}}$}\;

\end{algorithm}

\section{Large Language Model Prompting}
\label{app:llm_prompting}

We adopt the prompt design of~\citet{GLANCE}, which constructs prompts with different neighborhood ranges for each node: 
\begin{itemize}
\item \textbf{ego prompt}: contains only the textual description of the target node.
\item \textbf{one-hop prompt}:  contains the textual descriptions of the target node and its sampled one-hop neighbors.
\item \textbf{two-hop prompt}: contains the textual descriptions of the target node and its sampled two-hop neighbors, excluding the one-hop neighbors.
\end{itemize}
Each prompt begins with a node classification instruction and candidate categories. The target-node and neighborhood texts are then organized into labeled fields. The category names provide information about the prediction space, while the field labels distinguish the target node from its structural context. The three prompt templates are shown below:

\noindent{\textbf{Ego Prompt example}:}\par

\begin{lstlisting}[basicstyle=\ttfamily\small, breaklines=true]
Instruct: Predict the node's category from the provided
ego-node context. Possible categories: <candidate categories>.
Query:
EGO:
<target-node text>
Category?
</END>
\end{lstlisting}

\noindent{\textbf{One-hop Prompt example}:}\par

\begin{lstlisting}[basicstyle=\ttfamily\small, breaklines=true]
Instruct: Predict the node's category from the provided
1-hop context. Possible categories: <candidate categories>.
Query:
EGO:
<target-node text>
HOP1:
- <one-hop neighbor 1 text>
- <one-hop neighbor 2 text>
...
Category?
</END>
\end{lstlisting}

\noindent{\textbf{Two-hop Prompt example}:}\par

\begin{lstlisting}[basicstyle=\ttfamily\small, breaklines=true]
Instruct: Predict the node's category from the provided
2-hop context. Possible categories: <candidate categories>.
Query:
EGO:
<target-node text>
HOP2:
- <two-hop neighbor 1 text>
- <two-hop neighbor 2 text>
...
Category?
</END>
\end{lstlisting}

\section{Experimental Details}
\subsection{Introduction of Datasets}
\label{app:node_dataset}

To characterize the limitations of current methods, as well as evaluate DeKoR, 
we use five node classification datasets: 
Cora~\citep{CoraCiteseer}, 
CiteSeer~\citep{CoraCiteseer}, 
PubMed~\citep{Pubmed}, 
WikiCS~\citep{Wikics}, and 
arXiv~\citep{arXiv}. 
Table~\ref{tab:dataset_statistics} presents the statistical information 
for the five graph datasets.

\begin{itemize}
\item \textbf{Cora.}
Cora is a citation network of scientific publications in the field of machine learning. 
Each node is a paper, and an edge indicates a citation relationship between two papers. Each paper is described by a bag-of-words feature vector. The nodes are categorized into 7 different AI fields.
\item \textbf{CiteSeer.}
CiteSeer is a citation network containing scientific publications from multiple research areas. Nodes correspond to publications, while edges represent citation relationships between papers. Each node is associated with a bag-of-words feature vector derived from its text content. The node classification task involves 6 research categories.
\item \textbf{PubMed.}
PubMed is a larger citation network of biomedical publications, where papers have an
edge if they have a citation relationship. Each article is represented by a TF–IDF word vector, with labels corresponding to 3 different disease types.
\item \textbf{WikiCS.}
WikiCS is a text-attributed graph constructed from Wikipedia computer science articles. 
Each node corresponds to a computer science article, and edges capture relationships between related articles. Each node is represented by a 300-dimensional feature vector, and the nodes are categorized into 10 different computer science topics.
\item \textbf{arXiv.}
arXiv is a large-scale citation network of computer science papers collected from arXiv. Nodes represent scientific papers, and edges are citation relationships between papers. Each paper is associated with a 128-dimensional feature vector, and the node classification task covers 40 research categories.
\end{itemize}

\begin{table}[H]
\caption{Summary statistics of the datasets used in our experiments.}
\label{tab:dataset_statistics}
\vspace{\baselineskip}
\centering
\small
\setlength{\tabcolsep}{8pt}
\renewcommand{\arraystretch}{1.11}

\begin{tabular}{l|cccc}
\outerrule
\textbf{Dataset} &
\textbf{\#Nodes} &
\textbf{\#Edges} &
\textbf{Original Feature Dim.} &
\textbf{\#Classes} \\
\grouprule

Cora       & 2{,}708   & 5{,}429      & 1{,}433 & 7  \\
CiteSeer   & 3{,}186   & 4{,}277       & 3{,}703 & 6  \\
PubMed     & 19{,}717  & 44{,}335      & 500     & 3  \\
WikiCS     & 11{,}701  & 215{,}863     & 300     & 10 \\
arXiv      & 169{,}343 & 1{,}166{,}243  & 128     & 40 \\

\outerrule
\end{tabular}
\vspace{2pt}
\end{table}

\subsection{Construction of Subgraph Classification Datasets}
\label{app:subgraph_dataset}
To evaluate our method on graph-level tasks, we transform each node classification dataset into a subgraph classification dataset. Given a text-attributed graph, we construct a 2-hop subgraph centered at each node and use the center node's label as the subgraph label. For each center node $v_i$, we construct an adjacency list from the graph structure, symmetrize the adjacency list and remove self-loops. We then randomly sample at most five 1-hop neighbors of $v_i$. For each selected 1-hop neighbor, we further sample at most five candidate neighbors to form the 2-hop expansion set. During this process, the center node and selected 1-hop neighbors are excluded, and duplicate 2-hop nodes reached through different intermediate neighbors are removed. The 2-hop expansion set therefore contains at most $5\times5=25$ nodes, resulting in at most $1+5+25=31$ nodes in each sampled subgraph. Formally, the node set of the subgraph centered at $v_i$ is defined as:
\begin{equation}
    \mathcal{V}_i^{\mathrm{sub}}
    =
    \{v_i\}
    \cup
    \mathcal{N}_i^{(1)}
    \cup
    \mathcal{N}_i^{(2)},
\end{equation}
where $\mathcal{N}_i^{(1)}$ and $\mathcal{N}_i^{(2)}$ denote the sampled one-hop and two-hop neighbors. The subgraph edge set contains the original edges whose endpoints are both in $\mathcal{V}_i^{\mathrm{sub}}$. The nodes in each subgraph are relabeled with local indices, with the center node assigned index zero. The corresponding node features and labels are extracted from the original graph, and each node is additionally assigned a hop type of 0, 1, or 2 according to its distance from the center. We also retain the original node IDs and raw texts to align the subgraph nodes with the LLM teacher. Using a fixed random seed, we construct and store the resulting subgraphs as PyTorch Geometric \texttt{Data} objects.

\subsection{Introduction of Baselines}
\label{app:baselines}
\textbf{Graph Neural Networks (GNNs)}. The following outlines the GNN baselines used for comparison. 

\begin{itemize}
\item GCN~\citep{GCN}: GCN applies graph convolution to aggregate information from neighboring nodes, combining node features with local graph structure to generate node representations.

\item GAT~\citep{GAT}: GAT introduces an attention mechanism into graph convolution. It assigns different weights to neighboring nodes when aggregating their information.

\item GraphSAGE~\citep{GraphSAGE}: GraphSAGE generates node representations by sampling neighboring nodes and aggregating their features with a learnable aggregation function. It enables scalable graph representation learning.

\item GIN~\citep{GIN}: GIN combines a node's own features with aggregated neighbor information through a learnable multilayer perceptron to learn expressive node representations.
\end{itemize}

\textbf{Graph Transformers (GTs)}. We next introduce the GT baselines used for comparison. 

\begin{itemize}
\item DIFFormer~\citep{DIFFormer}: This GT architecture employs a diffusion-based attention mechanism to efficiently capture global dependencies between nodes.

\item NodeFormer~\citep{NodeFormer}: This GT architecture employs a kernelized Gumbel-softmax operator to efficiently learn graph structures in a differentiable manner.

\item DUALFormer~\citep{DUALFormer}: This GT architecture employs a dual-dimensional design to capture local inter-node dependencies and global inter-feature correlations efficiently.

\item CoBFormer~\citep{CoBFormer}: This GT architecture employs self-attention mechanisms both within and across clusters to mitigate the over-globalization problem.
\end{itemize}

\textbf{GNN--LLM Fusion Methods.} We next introduce the GNN--LLM fusion baselines used for comparison.

\begin{itemize}
\item UniGraph~\citep{UniGraph}: This framework builds a cross-domain graph foundation model by combining language models with GNNs, using masked graph modeling for pre-training and graph instruction tuning to support few-shot and zero-shot transfer.

\item OFA~\citep{OFA}: This framework unifies cross-domain graph data as text-attributed graphs, encodes node, edge, and task descriptions with language models, and uses nodes-of-interest and graph prompting to enable a single GNN to handle classification tasks.

\item GraphGPT~\citep{GraphGPT}: This framework integrates graph structural knowledge into LLMs through text-graph grounding, a lightweight graph-text alignment projector, and dual-stage graph instruction tuning.

\item LinguGKD~\citep{LinguGKD}: This framework uses instruction tuning to train an LLM teacher and adopts layer-adaptive contrastive knowledge distillation to transfer semantic and relational knowledge from the LLM to a lightweight GNN.

\item GLANCE~\citep{GLANCE}: This framework employs a lightweight router to selectively invoke the LLM for nodes where the GNN is likely to underperform, and then fuses the LLM embeddings with GNN representations for classification through a refiner.
\end{itemize}

For the four GNN baselines, i.e., GCN, GAT, GraphSAGE, and GIN, we use the public library PyTorch Geometric (PyG) ~\citep{PyG} for their implementation. For other baselines, we leverage their source codes. The sources are outlined as: 

\begin{itemize}[itemsep=0pt, topsep=0pt]
\raggedright
\item GNNs: \url{https://github.com/pyg-team/pytorch_geometric/tree/master/torch_geometric/nn/conv}
\item DIFFormer: \url{https://github.com/qitianwu/DIFFormer}
\item NodeFormer: \url{https://github.com/qitianwu/NodeFormer}
\item DUALFormer: \url{https://github.com/JiamingZhuo/DUALFormer}
\item CoBFormer: \url{https://github.com/null-xyj/CoBFormer}
\item UniGraph: \url{https://github.com/yf-he/UniGraph}
\item OFA: \url{https://github.com/LechengKong/OneForAll}
\item GraphGPT: \url{https://github.com/HKUDS/GraphGPT}
\item LinguGKD: \url{https://github.com/SXiangHu/LinguGKD}
\item GLANCE: \url{https://github.com/dloveland/GLANCE-ICLR26}
\end{itemize}

For all node classification and subgraph classification experiments, we preserve the original design of each baseline as much as possible. Specifically, we retain the original model architecture, graph pre-processing strategy, and training objective. For GNN--LLM methods, we also preserve the original prompt templates.

For subgraph classification experiments that require adapting a node-level method to graph-level prediction, we append a global mean-pooling readout after the final representation layer to obtain the graph-level representation. This readout operation is applied only at the final stage and does not modify the baseline backbone. For baselines that already provide a graph-level readout, we retain their original prediction mechanism.

\subsection{Pretrain the GNN Encoder}
\label{app:gnn_pretraining}
We pre-train a graph neural network encoder for each dataset to capture structural knowledge, using GCNII~\citep{GCNII} as the backbone. The GNN uses a hidden dimension of 64 and a dropout probability of 0.2. For data splitting, we use an independent random seed of 123 to partition the nodes into training, validation, and test sets with a ratio of 50\%/25\%/25\%, respectively. The GNN trainer uses at most 50,000 training nodes and 50,000 validation nodes. We optimize the GNN parameters using AdamW with a learning rate of $1\times10^{-3}$ and a weight decay of $1\times10^{-4}$. The maximum number of training epochs is set to 200, with an early-stopping patience of 50 epochs.

\subsection{Fine-tuning the LLM}
\label{app:llm_finetuning}
We separately fine-tune a pre-trained LLM for each dataset to produce semantic representations. We adopt Qwen-8B with LoRA-based parameter-efficient fine-tuning, where the LoRA rank is set to 8. The LLM is optimized using the AdamW optimizer with a learning rate of $1\times10^{-4}$. The maximum number of training epochs is set to 10, with an early-stopping patience of 2 epochs. The LLM trainer follows the same deterministic data-splitting scheme as the GNN trainer, partitioning the nodes into training, validation, and test sets with a ratio of 50\%/25\%/25\%, respectively, using a default random seed of 123. To reduce the computational cost of fine-tuning, we limit the training and validation sets to at most 3,000 and 1,000 nodes, respectively.

The LLM is trained using three prompts described in Appendix~\ref{app:llm_prompting}: the ego prompt, one-hop prompt, and two-hop prompt. We use left padding and left truncation during tokenization. The maximum sequence length is 1,024 tokens for the ego prompt and 4,096 tokens for one-hop and two-hop prompts. For each prompt, the hidden state corresponding to the last non-padding token is used as the pooled representation. We concatenate the representations from the three prompts and use an MLP to map the embeddings to the label space for each dataset. After fine-tuning, we remove the MLP head and use the fine-tuned LLM to generate semantic representations for our method.

\subsection{Node and Subgraph Classification Experiments}

For the final node and subgraph classification experiments, we use a 50\%/25\%/25\% split for the training, validation, and test sets, respectively. To reduce the computational cost on large datasets, we limit the numbers of training, validation, and test nodes to 3,000, 1,000, and 10,000, respectively. For the node and subgraph classification baselines, we preserve the original hyperparameter settings and only tune the learning rate over ${0.01, 0.001, 0.0001, 0.00001}$. For our method, in addition to tuning the learning rates of the expert distillation and routing, we search $\lambda_{\mathrm{loc}}, \lambda_{\mathrm{glo}}$ and $\tau_{\mathrm{loc}}, \tau_{\mathrm{glo}}$ over ${0.1, 1, 10, 100}$, and the number of prototypes $K$ over ${8, 16, 32, 64}$. All experiments are conducted with three random seeds, 123, 456, and 789, and we report the mean accuracy and standard deviation over three runs.

\end{document}